\def\year{2019}\relax
\documentclass[letterpaper]{article} 
\usepackage{aaai19}  
\usepackage{times}  
\usepackage{helvet}  
\usepackage{courier}  
\usepackage{url}  
\usepackage{graphicx}  

\usepackage{epstopdf}
\usepackage{algorithm}
\usepackage{algorithmicx}
\usepackage{algpseudocode}
\usepackage{graphics}
\usepackage{epsfig}
\usepackage{mathptmx}
\usepackage{amsmath}
\usepackage{amssymb}
\usepackage{amsthm}
\usepackage{natbib}
\usepackage{siunitx}
\usepackage{multirow}
\usepackage{color}
\begin{document}
%
\title{Deep Feature Learning of Multi-Network Topology for Node Classification}

\author{Hansheng Xue, \textsuperscript{1,2} Jiajie Peng, \textsuperscript{1}\thanks{Corresponding author}, Xuequn Shang\textsuperscript{1}\\
\textsuperscript{1}{School of Computer Science, Northwestern Polytechnical University, Xi’an, China}\\
\textsuperscript{2}{School of Computer Science and Technology, Harbin Institute of Technology, Shenzhen, China}\\
xhs1892@gmail.com, jiajiepeng@nwpu.edu.cn, shang@nwpu.edu.cn\\
}

\maketitle
\begin{abstract}
Networks are ubiquitous structure that describes complex relationships between different entities in the real world. 
%
As a critical component of prediction task over nodes in networks, learning the feature representation of nodes has become one of the most active areas recently.
Network Embedding, aiming to learn non-linear and low-dimensional feature representation based on network topology, has been proved to be helpful on tasks of network analysis, especially node classification. For many real-world systems, multiple types of relations are naturally represented by multiple networks. However, existing network embedding methods mainly focus on single network embedding and neglect the information shared among different networks. In this paper, we propose a novel multiple network embedding method based on semi-supervised autoencoder, named DeepMNE, which captures complex topological structures of multi-networks and takes the correlation among multi-networks into account. We evaluate DeepMNE on the task of node classification with two real-world datasets. The experimental results demonstrate the superior performance of our method over four state-of-the-art algorithms.

\end{abstract}

\section{Introduction}
Networks are powerful sources for describing and modeling complex systems. Mining knowledge from networks has become a popular yet challenging area. Many researchers have started to focus on this area. One of the most important tasks in network analysis is node classification. In a typical node classification task, the aim is to predict the most probable labels for nodes in a given network~\citep{tsoumakas2007multi}.
For example, in a protein-protein interaction network, the aim is to predict functional labels of proteins~\citep{radivojac2013large}.


A batch of informative features is required and important for a supervised machine learning method~\citep{grover2016node2vec}. Node classification problem is always solved as a supervised machine learning problem. Therefore, a feature vector representation for nodes in the network should be appropriately constructed.
 Recently, learning the feature representation of a node based on its neighbors and structural information of the network has become one of the most active areas~\citep{grover2016node2vec,Perozzi2014DeepWalk,Zhang2015LINE}. Learning low-dimensional feature representation of nodes is also termed as $Network\ Embedding$, which has recently attracted lots of attentions~\citep{Goodfellow2016}. Besides node classification~\citep{Jian2018Toward}, the outputs of network embedding have also been widely used on many important tasks, such as link prediction~\citep{Li2018Streaming} and community detection~\citep{Yang2013Community}.

For many real-world systems, multiple types of relations are naturally represented by multiple networks. For example, in social network, the relations between people include friendship relation, money transferring relation, colleague relation and so on. Since multiple networks can describe the real-world systems better in many cases, multi-networks analysis have attracted lots of attention in network science community recently.
Unfortunately, most existing network embedding methods focus on single network, and few approaches concentrate on learning node representation based on multiple networks. Therefore, it is urgent to develop an algorithm for learning the feature representation of a node by integrating multi-networks appropriately.
%


%
%

One simple solution for multi-network embedding is to summarize multiple networks into a single network and apply the single-network embedding method on the integrated network. Several multi-networks integration methods have been proposed, such as probabilistic methods~\citep{Franceschini2013STRING}, kernel-based methods~\citep{Yu2015Predicting} or weighted averaging or summing~\citep{Mostafavi2008GeneMANIA}.
However, this type of integration methods often result in information loss problem when integrating multiple networks into a single one~\citep{Tsuda2005Fast,Lanckriet2004A}. Some approaches try to train individual classifiers on different networks and combine these predictions to a final result using ensemble learning methods~\citep{Yan2010A}. However, these methods consider different networks as independent ones, ignoring the correlation between different networks. In addition, such methods often suffer from learning time and memory constraints~\citep{Gligorijevic2017deepNF}.



In multi-network embedding, multiple networks represent different types of relations among the same set of nodes (representing person, commodity, gene and so on). There may be potential correlations between different networks. For multi-networks embedding, one of the most challenging task is how to consider the correlation between different networks. To address this problem, we try to model the correlation between different networks during the feature learning process.

Autoencoder~\citep{Rumelhart1986Learning, Baldi2011Autoencoders} is a typical unsupervised deep learning model, which aims to learning a new encoding representation of input data.
It has been proved that autoencoder can solve these non-linear feature learning problems effectively. However, existing autoencoder-based methods are not designed for learning multi-network topological features. To benefit from the feature learning power of autoencoder and consider the correlation between multiple networks, we propose a novel multi-network-based feature learning algorithm, named DeepMNE. Considering correlation between multiple networks, DeepMNE applies stacked semi-autoencoder to map input multi-networks into a low-dimension and non-linear space. Here are the major contributions:

\begin{itemize}
\item We propose a novel semi-supervised autoencoder model for learning feature representations of nodes based on multiple networks.  

\item To consider the correlation between different networks, we design a communication mechanism among multiple autoencoders corresponding to multiple networks.

\item We empirically evaluate DeepMNE for multi-label classification on two tasks of gene function prediction. The experimental results show that DeepMNE outperforms the existing state-of-the-art methods.

\end{itemize}

\section{Related Work}
\subsection{Multi-network Embedding}
As an extension of single network embedding, multi-network embedding aims to represent nodes using low-dimensional topological information from multi-networks. Current network embedding approaches mainly focus on single-network embedding and utilize topological structure information to represent nodes.
For instance, DeepWalk~\citep{Perozzi2014DeepWalk} treats nodes as words and generates short random walks as sentences. Then, it uses Skip-gram, a word representation learning model, on these random walks to represent nodes of networks.
Similar, node2vec~\citep{grover2016node2vec} utilizes a biased random walking procedure to learn topological information, and it uses negative sampling to optimize the Skip-gram model. 
DNGR~\citep{Cao2016Deep} is a novel method which uses random surfing to learn topological information and applies stacked denoising autoencoder to generate low-dimensional node representation.
In general, all these methods focus on single-network representation learning, and few can apply on multi-networks directly.

Besides, some multi-network integration methods have been proposed in biological networks area. Mashup~\citep{Cho2016Compact} is an integrative framework for learning low-dimensional feature representations of genes form multiple networks constructed from heterogeneous data sources.
%
Similarity Network Fusion~\citep{Wang2014Similarity} is a widely used networks integration method, which constructs networks for each available data type and then efficiently fuses these networks into one.
%
In addition, there are some other multi-network integration methods, such as Diffusion State Distance~\citep{Cao2014New} and Collective Matrix Factorization~\citep{Z2015Gene}. However, these methods are linear and shallow approaches which cannot capture complex and highly non-linear structure across all networks. 

\subsection{Gene function prediction}
Accurate annotation of gene function is one of the most important and challenging problems in biological area. Predicting gene function aims to assign an unknown gene to the correct functional categories in the annotation database, such as Gene Ontology.
To solve this problem, lots of methods based on different types of biological information have been proposed, such as amino acid sequence-based method~\citep{Clark2011Analysis}, protein structure-based method~\citep{Pal2005Inference} and gene expression-based method~\citep{Huttenhower2006A}.
With the improvement of experimental methods, functional associations between genes or proteins are often represented in terms of networks, such as gene co-expression networks and protein-protein interaction networks. Several network-based gene or protein function prediction methods have been proposed~\citep{Lehtinen2015Gene,Sharan2007Network}. Multi-networks-based function predictions have been proved better than those methods based on single data source~\citep{Re2010Integration,Cozzetto2013Protein}, because of the complementary nature of different data sources. Thus, lots of algorithms have been proposed for gene function prediction by integrating multiple biological networks~\citep{Cho2016Compact,Sara2010Fast,Cao2014New}.

\section{Our Proposed Approach}
Multiple-network embedding can be formulated as a semi-supervised feature learning problem. In this part, we propose a novel semi-supervised autoencoder, termed as DeepMNE, to learn the node representation based on multi-networks.

Let $V$ be a set of $n$ nodes $\{v_1,v_2,...,v_n\}$.
Let $E$ be a set if edges between pairs of $n$-nodes $\{v_1, v_2,...,v_n\}$.
Given $k$ networks that include the same set of nodes $V$ but different connectivity between nodes, labeled as $\{G^{(1)}, G^{(2)}, ..., G^{(k)}\}$, a network $G_i$, each network is represented as $G^{(i)}=(V,E^{(i)})$, where $i \in \{1,2,...,k\}$.
Our aim is to learn a low-dimension feature representation for each $v \in V$ based on the topological information contained in $\{G^{(1)}, G^{(2)}, ..., G^{(k)}\}$. DeepMNE contains two main components: obtaining global structure information of each network; learning feature representation of nodes by considering both topology of multiple networks and their correlation.


\subsection{Step 1. Obtaining global structure information using RWR}
It has been proved that random walk with restart (RWR) could capture global associations between nodes in a network~\citep{Cho2016Compact}. Instead of inputting adjacency matrices into DeepMNE directly, we run RWR on each network to capture single network topological information and convert it into feature representations of nodes. The adjacency matrix only describes the relationships between any directly connected nodes, ignoring the global structure of a network. RWR can overcome this drawback, and represent nodes using these high-dimensional network structural information.
Besides, we choose the RWR method instead of other recently proposed network embedding methods, such as node2vec~\citep{grover2016node2vec} and DeepWalk~\citep{Perozzi2014DeepWalk}, to capture the topological information, because these methods are computationally intense and require additional hyper-parameter fitting~\citep{Gligorijevic2017deepNF}.

Let $M_k$ denote the adjacency matrix of a network $G^{(k)}=(V,E^{(k)})$.
The RWR from node $v_i$ can be described as the following recurrence relation.

\begin{equation}
s_i^{t+1}=(1-\alpha)Ts_i^t + \alpha e_i\label{eq:01}
\end{equation}

\noindent where $\alpha$ is the restart probability, which balances the effect of local and global topological information in the network; $e_i$ is a n-dimensional distribution vector with $e_i(i)=1$ and $e_i(j)=0$, $\forall j \neq i$; $s_i^t$ is a n-dimensional distribution (column) vector in which each entry holds the probability of a node being visited after $t$ steps in the random walk, starting from node $v_i$; $T$ is the transition probability matrix, and each entry $T_{ij}$, which stores the probability from node $j$ to node $v_i$, can be calculated as $T_{ij} = \frac{M_{ij}}{\sum_{i}{M_{ij}}}$. Based on RWR, we can obtain a matrix $S$, in which $S_{ij}$ is the relevance score between node $v_i$ and $v_j$ defined by RWR-based steady state probabilities.

\begin{figure*}[!tpb]
\centerline{\includegraphics[scale=0.061]{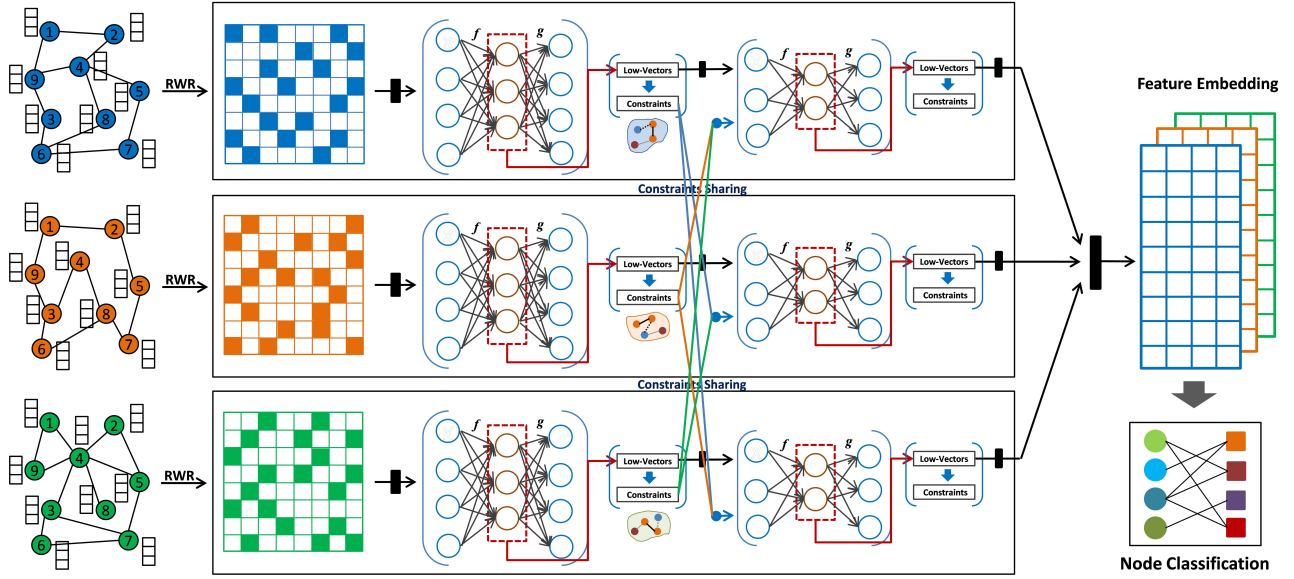}}
\caption{The structure of DeepMNE algorithm. The whole process mainly contains two parts, obtaining topological information learning and learning multi-network-based features. The output of DeepMNE could feed to the following machine learning model. We firstly run random walk with restart (RWR) to learn global structure of networks. Then, constraints extraction and application with semi-supervised autoencoder are iteratively implemented on DeepMNE algorithm to integrate multi-networks. After obtaining the integrated representations of multi-networks, we can train machine learning model based on the outputs of DeepMNE to classify nodes.}\label{fig:framework}
\end{figure*}

\subsection{Step 2. Multi-network embedding with semiAE}
In this section, we propose a novel multi-network embedding algorithm, termed as DeepMNE. The main framework is a DNN structure with autoencoder (AE) and Semi-Supervised autoencoder (semiAE) as its building block. The whole process includes two parts: constraints extraction and constraints application. We use constraints to capture the correlation between different networks. Given several networks $\{G^{(1)}, G^{(2)},..., G^{(k)} \}$ with same nodes, the input of this step is $\{S^{(1)}, S^{(2)},..., S^{(k)} \}$ calculated based on RWR. The main framework is shown in Figure~1\vphantom{\ref{fig:01}}.

The first layer of DeepMNE framework is the original autoencoder, which is used for feature extraction and dimension reduction. Starting from the second layer, a revised autoencoder (semiAE) is used for constraint integration and dimension reduction. The dimension of input networks decreases constantly with the extension of the whole iteration model.

\subsubsection{Prior Constraints Extraction}
%
The idea of constraints comes from semi-supervised clustering. The pairwise constraints can be typically formatted as must-link and cannot-link constraints~\citep{Basu2004A}. The pairwise constraints can be described as follows:
 a must-link constraint indicates that these nodes are highly similar or belong to the same cluster, while a cannot-link constraint indicates that two points in the pair are highly dissimilar or belong to different clusters.

Given pairs of nodes, we use two strategies to extract constraints. One is to calculate and sort pairwise pearson correlation coefficient (PCC) of all pairs of nodes based on their feature vectors. The top-$k$ pairs are considered as the must-link constraints and the bottom-$k$ pairs are considered as the cannot-link constraints. The other is to set two thresholds for must-link and cannot-link, labeled as $f_1$ and $f_2$ respectively. In detail, a pairs can be adopted as a must-link constraint if its PCC value is larger than $f_1$, and a pair is considered as cannot-link constraint if the PCC value is smaller than $f_2$.

After extracting the constraints from the previous layer ($i$ layer), we can apply the must-link and cannot-link constraints to the next layer ($i$+1 layer) as the prior information.

\subsubsection{Novel autoencoder with constraints}
The key question of DeepMNE is how to integrate prior constraints into the network representation through autoencoder. We revise the original autoencoder and propose a novel variant of autoencoder, termed as Semi-Supervised AutoEncoder (semiAE).
Starting from the second layer, the input includes both low-dimensional representations and constrains from previous layer. It is noted that constraints from previous layers' building blocks are based on different networks. Therefore, constraints may be conflicting. To solve this problem, we would merge these constraints and take the intersection of all the constraints as the input of semiAE.

Autoencoder is an unsupervised model which is composed of two parts, i.e. the encoder and decoder. The encoder transform the high-dimensional data into a low-dimensional code, and a similar "decoder" network to recover the data from the low dimensional code. The low-dimensional code is then used as a compressed representation of the original data.
%
Let $x_i$ be the $i$-th input vector or node representation of network, and $f$ and $g$ be the activations of the hidden layer and the output layer respectively. We have $h_i=f(Wx_i+b)$ and $y_i=g(Mh_i+d)$, where $\Theta=\{\theta_1,\theta_2\}=\{W,b,M,d\}$ are the parameters to be learned, $f$ and $g$ are the non-linear operators such as the sigmoid function ($sigmoid(z)=1/(1+exp(-z))$) or tanh function ($tanh(z)=(e^z-e^{-z})/(e^z+e^{-z})$). Then the optimization goal is to minimize the reconstruction error between the original data $x_i$ and the reconstructed data $y_i$ from the new representation $h_i$.

\begin{equation}
\arg\min_{\theta\in{\Theta}}\sum_{i=1}^{n}\parallel y_i-x_i\parallel^2 \label{eq:04}
\end{equation}

The original autoencoder cannot model the constraints obtained from previous layers. We propose semiAE to take these constraints into account.
Let $M$ be a set of must-link pairwise constraints where $(x_i,x_j)\in M$ implies the strong association between $x_i$ and $x_j$.  
Let $C$ be a set of cannot-link pairwise constraints where $(x_i, x_j)\in C$ implies $x_i$ and $x_j$ are unrelated. The number of constraints is much less than the size of the network $|M| + |C| \leq |S|$.

The hypothesis is that $x_i$ and $x_j$ should also close based on the low-dimensional space if there is a must-link constraint between them in previous layer.  Ideally,  after encoding, two must-link nodes should be closer, and two cannot-link nodes may be more distant.  
Mathematically, let $d(h(x_i),h(x_j))$ be the error score (difference) between $x_i$ and $x_j$ in the encoded space. For Must-link, $d(x_i,x_j)$ should be larger than $d(h(x_i),h(x_j))$; for Cannot-link, $d(x_i,x_j)$ should be smaller than $d(h(x_i),h(x_j))$.

If the pair $(x_i,x_j)$ is a must-link constraint, we add a penalty on the loss function. Similarity, if the pair $(x_i, x_j)$ is a cannot-link constraint, we add a reward on the loss function. The loss function for modeling constraints is defined as follows:

\begin{equation}
\begin{split}
& L_{mc}= \lambda_1\sum_{(x_i,x_j) \in M}d(h(x_i),h(x_j)) - \lambda_2\sum_{(x_i,x_j) \in C}d(h(x_i),h(x_j))\\
& =\lambda_1\sum_{i,j=1}^nM_{i,j}||h(x_i),h(x_j)||_2^2 - \lambda_2\sum_{i,j=1}^nC_{i,j}||h(x_i),h(x_j)||_2^2 \label{eq:05}
\end{split}
\end{equation}

\noindent where matrix $M$ and $C$ are set of must-link and cannot-link constraints respectively; $h(x_i)$ and $h(x_j)$ are the hidden layer representation of input feature vectors $x_i$ and $x_j$ that are from previous layer; $\lambda_1$ and $\lambda_2$ are the weight coefficient, controlling the influence of penalty and reward respectively.

To combine constraints with autoencoder, we propose a novel semi-supervised autoencoder, which integrates Eq.~(\ref{eq:04}) and Eq.~(\ref{eq:05}) and joint minimizes the following objective function:

\begin{equation}
loss = \arg\min_{\theta\in{\Theta}}\sum_{i=1}^{n}\parallel y_i-x_i\parallel^2 + \lambda L_{mc}\label{eq:06}
\end{equation}

\noindent The first part of Equation~\ref{eq:06} measures the squared error between input and output node features, and the second part measures error score of constraints in hidden layer.

\subsubsection{The DeepMNE multi-networks integration algorithm}
To optimize the aforementioned model, the goal is to minimize the loss function Eq.~(\ref{eq:06}). In detail, the key step is to calculate the partial derivative of $\frac{\partial L_{mc}}{\partial W}$. And the loss function of $L_{mc}$ can be rephrased as follows:

\begin{equation}
\begin{split}
& L_{mc}= \lambda_1\sum_{i,j=1}^nM_{i,j}||h(x_i),h(x_j)||_2^2 - \lambda_2\sum_{i,j=1}^nC_{i,j}||h(x_i),h(x_j)||_2^2\\
& = 2\lambda_1tr(H^\mathrm{ T }L_MH)-2\lambda_2tr(H^\mathrm{ T }L_C H) \\
& = tr(H^\mathrm{ T }(L_M-L_C)H) \label{eq:07}
\end{split}
\end{equation}

\noindent where $L_M = D_M - M$, $D_M\in{\mathbb{R}^{n\times n}}$ is a diagonal matrix, $D_{M_{i,j}}$ = $\sum_jM_{i,j}$. And $L_C$ is similar as $L_M$. $H$ is the simplified representation of hidden layer. Thus, $\frac{\partial L_{mc}}{\partial W}$ can be translated as:

\begin{equation}
\frac{\partial L_{mc}}{\partial W} = \frac{\partial L_{mc}}{\partial H} \centerdot \frac{\partial H}{\partial W}=\frac{\partial tr(H^\mathrm{ T }(L_M-L_C)H)}{\partial H} \centerdot \frac{\partial f(XW+b)}{\partial W} \label{eq:08}
\end{equation}

\noindent where $f$ is activation function(i.e. sigmoid), and we can obtain the partial derivatives of $L_{MC}$. Thus, with an initialization of the parameters, the novel semiAE can be optimized by using stochastic gradient descent (SGD).

The pseudocode for DeepMNE is given in Algorithm 1.

\begin{algorithm}
\caption{The DeepMNE algorithm}
\begin{algorithmic}[1]
\renewcommand{\algorithmicrequire}{\textbf{Input:}}
\renewcommand{\algorithmicensure}{\textbf{Output:}}
\Require Multi-networks $G=\{G^{(1)}, G^{(2)},...,G^{(K)}\}$ with $G^{(i)}=(V,E^{(i)})$, the number of iteration $T$, the percentage of constraints $P$, initialization parameters;
\Ensure Feature representation of nodes in $V$;
\State Run Random Walk with Restart on multi-networks $G$;
\State Train AutoEncoder to obtain novel feature representations of nodes in$G'$ and extract initial must-link, cannot-link constraints $M$, $C$;
\ForAll{$i \in T$}
\ForAll{$k \in K$}
\State $M'$, $C'$ = Merge constraints from other networks $M_{all \neq k}$, $C_{all \neq k}$;
\State $G_k'$ = Train semi-AutoEncoder with initial parameters on $G_k$ to optimize Eq.~(\ref{eq:06});
\State $M_k$, $C_k$ = Extract must-link and cannot-link constraints based on $G_k'$;
\EndFor
\EndFor \\
\Return Feature representation of nodes in $V$
\end{algorithmic}
\end{algorithm}

In the first phase of DeepMNE algorithm, we run random walk with restart algorithm to learn global structure of single biological network. Then, we train autoencoder to learn low-dimensional feature and extract prior constraints based on the network representation of hidden layer.
%
In the main phase, DeepMNE algorithm use an iterative model to train semi-supervised autoencoder with prior constraints. In each iteration, DeepMNE mainly contains three steps, which are merging constraints, training semi-autoencoder and extracting novel constraints. With the increasing of iterations, the model trend to converge and the constraints trends to be unchanged. Finally, DeepMNE generates several low-dimensional feature representations of nodes. 

The DeepMNE algorithm is a scalable framework model, its training complexity is linear to the number of vertexes $N$. The part of extracting constraints need to calculate pair-wise PCC value which requires $O(N^2)$ . Therefore, the training complexity of DeepMNE algorithm is $O((N^2 + N)TK)$, where $T$ is the number of iteration and $K$ is the number of multiple-networks.


\section{Experiments}
In order to evaluate the performance of DeepMNE, we test our method on a task of gene function prediction. Gene function prediction is a multi-label classification problem, which aims to assign unknown genes to the correct functional categories in the annotation database~\citep{Cho2016Compact}. 

We compare our model with the four state-of-the-art network embedding methods (Mashup~\citep{Cho2016Compact}, SNF~~\citep{Wang2014Similarity}, node2vec~\citep{grover2016node2vec} and DeepWalk~\citep{Perozzi2014DeepWalk}) and apply the integrated outputs on the task of gene function prediction using support vector machine. We adopt accuracy, micro-averaged F1, micro-averaged area under precision-recall curve (micro-AUPRC) and micro-averaged area under receiver operating characteristic curve (micro-AUROC) as evaluation metrics. We adopt 5-fold cross-validation to evaluate the performance.


\subsection{Datasets}
In the experimental part,  we evaluate the performance of our method on datasets of Yeast and Human respectively, which collected from the STRING database v9.1~\citep{Franceschini2013STRING}. 

\begin{itemize}
  \item \textbf{Yeast} - consisted of six networks over 6,400 genes. The detailed number of edges are listed on the Table 1, where each edge represents the probability of edge presence and the weight between 0 and 1. The functional labels are obtained from Munich Information Center for Protein Sequences~\citep{Ruepp2004The}. The functional categories in MIPS are organized in a three-layered hierarchy, and the number of functional categories in each layer are listed on the Table 2.


  \item \textbf{Human} - consisted of six networks over 18,362 genes with the number of edges varying from 3,760 to 1,576,332, and the value of every edge are between 0 and 1 (see Table 1). The functional labels are downloaded from the Gene Ontology database~\citep{Ashburner2000Gene}. We group the GO terms for human to obtain three distinct levels of functional categories for different specificities. The details are listed in Table 2.
\end{itemize}

\begin{table}[ht]
\caption{Statistics of datasets. The number of edges in different networks and the average degree of nodes $\langle k\rangle$.} 
\centering 
\begin{tabular}{c c c c c}
\hline\hline
\multirow{2}{*}{Network type} & \multicolumn{2}{c}{Yeast} & \multicolumn{2}{c}{Human}\\
~ & \# edges & $\langle k\rangle$ & \# edges & $\langle k\rangle$\\
\hline 
co-expression & 314,013 & 98.129 & 1,576,332 & 171.695 \\
cooccurence & 2,664 & 0.833 & 36,128 & 3.935 \\
database & 33,486 & 10.464 & 319,004 & 34.746 \\
experimental & 219,995 & 68.748 & 618,574 & 67.375 \\
fusion & 1,361 & 0.425 & 3,760 & 0.410 \\
neighborhood & 45,610 & 14.253 & 104,958 & 11.432 \\ 
\hline 
\end{tabular}
\label{table:01} 
\end{table}

\begin{table}[ht]
\caption{Function statistics. The whole dataset contains twelve mini-datasets with different numbers of functions.} 
\centering 
\begin{tabular}{c c c c c}
\hline\hline
\multicolumn{2}{c}{Dataset} & \multicolumn{3}{c}{\# numbers of labels} \\
\hline 
\multirow{3}{*}{Yeast} & level-1 & \multicolumn{3}{c}{17} \\
~ & level-2 & \multicolumn{3}{c}{74} \\
~ & level-3 & \multicolumn{3}{c}{154} \\
\hline
\multirow{4}{*}{Human} & & BP & CC & MF\\
\cline{2-5}
~ & 11-30 & 262 & 82 & 153 \\
~ & 31-100 & 100 & 46 & 72 \\
~ & 101-300 & 28 & 20 & 18 \\ 
\hline 
\end{tabular}
\label{table:02} 
\end{table}

\subsection{Parameter Settings}
The parameters vary with different datasets. The dimension of each layer on multi-networks integration framework (DeepMNE) is listed in Table 3.

In our model, we used restart probability of 0.5 for RWR, which is same as Mashup. The final dimension of network representation are 500 and 800 respectively. The whole DeepMNE-based multi-networks integration algorithm is optimized by using stochastic gradient descent~\citep{Bottou1991Stochastic}. The batch size is 128, the initial learning rate is 0.1 for yeast and 0.2 for human, and the epochs are 200 and 400 respectively.
For SNF, we generate an emsemble network and we run singular value decomposition to learn low-dimensional feature representation, and the dimension is same with our model. For node2vec and DeepWalk, we all use the default parameters. For all compared algorithms, we use SVM as the classifier to predict the function labels of genes.

\begin{table}[ht]
\caption{Neural Network Structures} 
\centering 
\begin{tabular}{c c} 
\hline\hline 
Datasets & \#nodes in each layer \\ [0.5ex] 
\hline 
Yeast & [6400-5220-4040-2860-1680-500] \\
Human & [18362-9181-4580-2295-1148-800] \\ [1ex] 
\hline 
\end{tabular}
\label{table:03} 
\end{table}

\subsection{Experimental Results on Yeast}
The gene function prediction mainly contains three parts: RWR-based global structure caption of single network, low-dimensional feature learning, and SVM-based gene function prediction.
In this section, we apply all compared approaches to predict functions of yeast genes based on six networks.  
All approaches are tested on three tasks corresponding to function labels at different levels (level 1 with 17 categories, level 2 with 74 categories and level 3 with 154 categories). The functional classification at level 1 is more general than level 2 and level 3. Similarly, The functional classification at level 2 is more general than level 3.

\begin{figure}[!tpb]
\centerline{\includegraphics[scale=0.225]{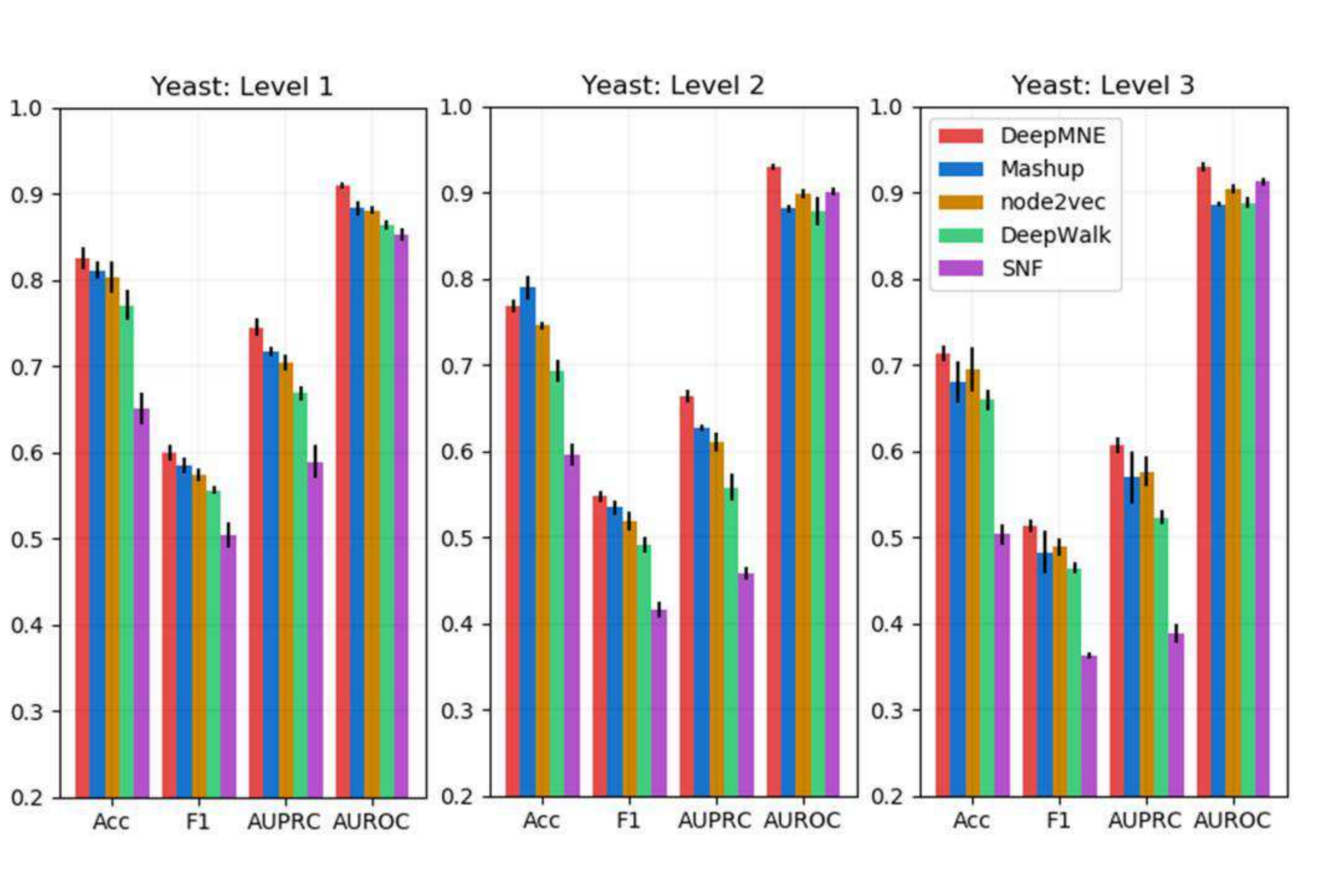}}
\caption{Performance comparison of different metrics on the task of predicting functional labels for yeast genes.}\label{fig:yeast_plot}
\end{figure}

Overall, comparing DeepMNE with four other approaches, DeepMNE can achieve better performance on yeast dataset at all three functional classification levels.
At level 1, DeepMNE can achieve the highest ACC score that is 0.8378\ (on average), in constract to 0.8063\ for Mashup and 0.6734\ for SNF. Besides, the micro-F1 score of DeepMNE algorithm is 0.7096, which is also higher than other four methods. The micro-average AUPRC and AUROC achieved by DeepMNE on level 1 of yeast are 0.7405 and 0.9100 respectively, which are significantly higher than the scores of Mashup and SNF (see Figure~\ref{fig:yeast_plot}).
The performance ranks of the four compared approaches are different at different levels. For example, Mashup is the second best approach at level 1, but node2vec goes to the second place at level 3. However, DeepMNE consistently achieve the best performance at different levels of labels.



\subsection{Experimental Results on Human}
For further evaluation, we also apply DeepMNE on human dataset to investigate its performance. Instead of RWR-based global structure information, we use the original adjacent matrix as the input of semi-autoencoder directly in this experiment. We test both types of input. Adjacent matrix can achieve better performance in this dataset, since human gene networks includes 18,362 nodes, which may be too large to capture the global structure.
As described on the previous section, nodes in human gene networks have three types of labels, termed biological process (BP) , cellular component (CC) and molecular function (MF),  corresponding to three respects of gene description.  Each type of labels are grouped to three levels, annotating 11-30, 31-100 and 101-300 genes respectively. Thus, we can obtain nine distinct mini datasets. We test all approaches on these datasets.

\begin{figure}[!tpb]
\centerline{\includegraphics[scale=0.225]{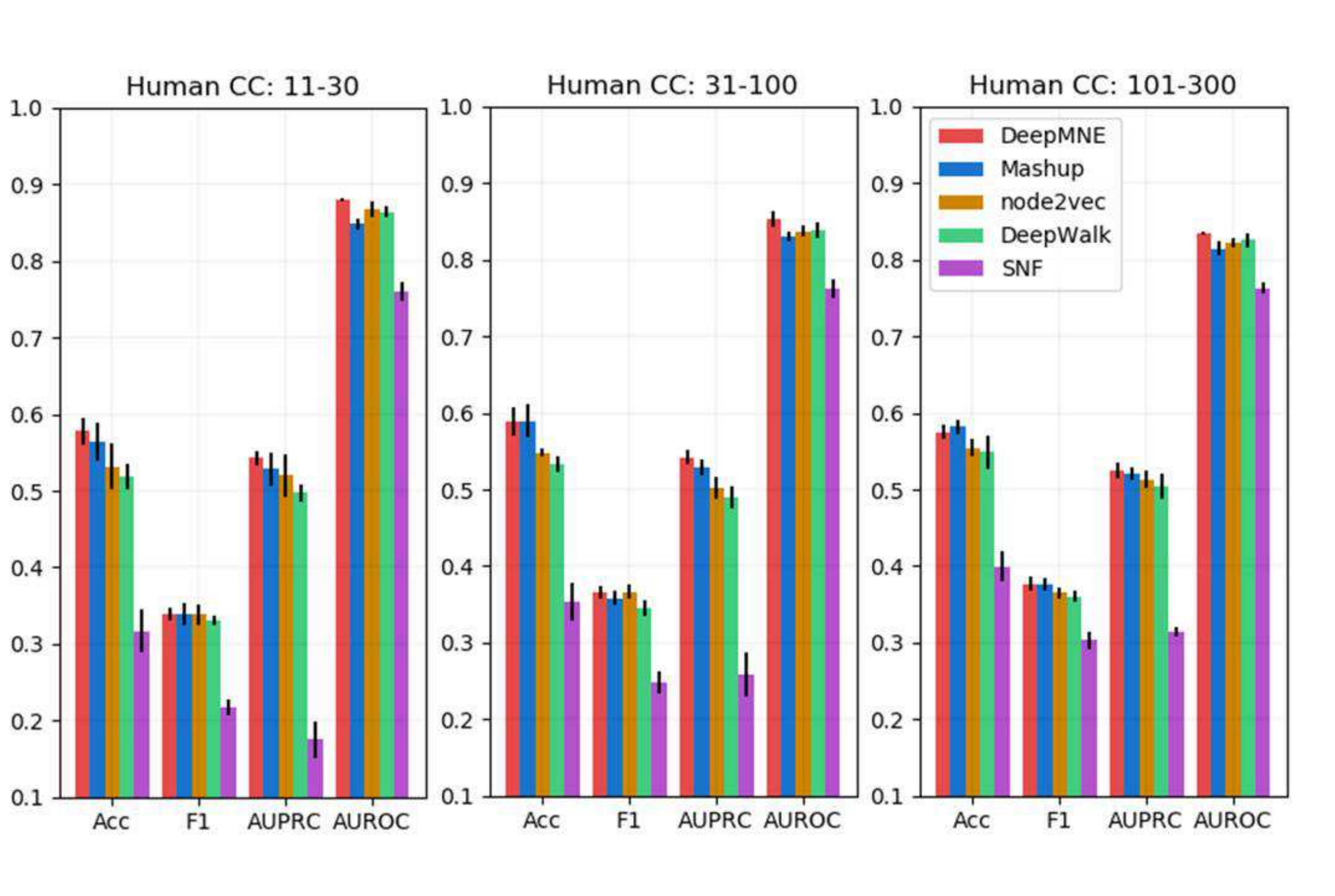}}
\caption{Performance comparison of different metrics on the task of predicting cellular component labels for human genes. }\label{fig:human_cc}
\end{figure}

Overall, DeepMNE also achieve the highest performance on human dataset(see Figure ~\ref{fig:human_cc}).
The accuracy of DeepMNE on human CC-11-30 is 0.5779, which is higher than Mashup, SNF, node2vec and DeepWalk (0.5644, 0.3167, 0.5480 and 0.5181 respectively).
DeepMNE still achieves the highest micro-F1 (0.3392), which is sightly higher than the other four methods (Mashup, SNF, node2vec, DeepWalk are 0.3388, 0.2176, 0.3383 and 0.3309 respectively).
The AUPRC values of DeepMNE implemented on three categories of human CC are 0.5430, 0.5418 and 0.5246, which are all higher than other four methods (0.5284, 0.5291 and 0.5202 for Mashup, 0.1852, 0.2588 and 0.3141 for SNF, 0.5272, 0.5242 and 0.5128 for node2vec, 0.4971, 4891 and 0.5043 for DeepWalk).
Besides, the AUROC values of DeepMNE (0.8796, 0.8533 and 0.8350 respectively) are all significantly higher than Mashup (0.8489, 0.8304, 0.8148), SNF (0.7602, 0.7623, 0.7633), node2vec (0.8675, 0.8376, 0.8227) and DeepWalk (0.8644, 0.8493, 0.8254).

The experimental results of Biological Process and Molecular Function categories are listed on Table 4. It is shown that DeepMNE outperforms other methods on MF
category of human dataset and also achieves good performance on BP categories.

\subsection{Parameters Analysis}

\begin{table}[!t]
\small
\caption{The Accuracy, AUPRC of DeepMNE on gene function prediction on human dataset.}\label{Tab:effects}
\centering
\renewcommand\arraystretch{1.1}
\setlength{\tabcolsep}{1.8mm}{
\begin{tabular}{cp{1.0cm}ccccc}
\hline
\hline
  &  & \multicolumn{2}{c}{\textbf{Biological Process}} & \multicolumn{2}{c}{\textbf{Molecular Function}}\\
  &  & Acc & AUPRC & Acc & AUPRC\\
\hline
  & Mashup & \textbf{0.3825} & \textbf{0.2318} & 0.4486 & 0.3836 \\
  & SNF & 0.1972 & 0.0602 & 0.3006 & 0.1662 \\
11-30 & node2vec & 0.3647 & 0.2248 & 0.4482 & 0.3742 \\
  & DeepWalk & 0.3613 & 0.2224 & 0.4466 & 0.3762 \\
  & DeepMNE & 0.3672 & 0.1910 & \textbf{0.4751} & \textbf{0.3897} \\
\hline
  & Mashup & 0.4113 & \textbf{0.2587} & 0.4717 & 0.3666 \\
  & SNF & 0.2376 & 0.0892 & 0.2689 & 0.1546  \\
31-100 & node2vec & 0.3812 & 0.2454 & 0.4355 & 0.3456 \\
  & DeepWalk & 0.3852 & 0.2477 & 0.4488 & 0.3654 \\
  & DeepMNE & \textbf{0.4129} & 0.2459 & \textbf{0.4936} & \textbf{0.4002} \\
\hline
  & Mashup & 0.4809 & 0.3795 & 0.5761 & 0.5236 \\
  & SNF & 0.3374 & 0.2023 & 0.4248 & 0.3473 \\
101-300 & node2vec & 0.4721 & 0.3740 & 0.3782 & 0.4959 \\
  & DeepWalk & 0.4802 & \textbf{0.3836} & 0.5365 & 0.5011 \\
  & DeepMNE & \textbf{0.4824} & 0.3692 & \textbf{0.5882} & \textbf{0.5406}\\
\hline
\end{tabular}}
\end{table}

\begin{figure}[!tpb]
\centerline{\includegraphics[scale=0.24]{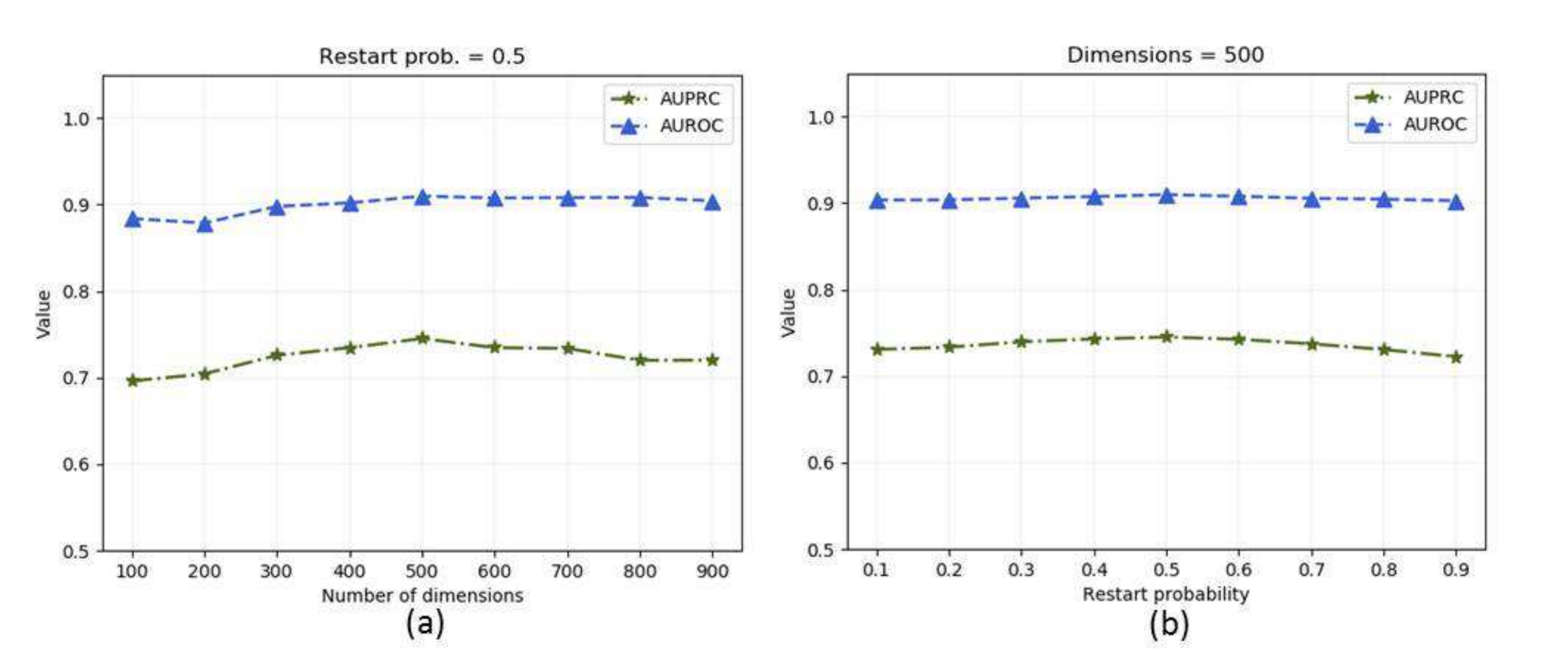}}
\caption{The AUPRC and AUROC score of DeepMNE with different restart probabilities and numbers of dimensions for function prediction on yeast dataset.}\label{fig:robustness}
\end{figure}

To evaluate the effect of restart probability to DeepMNE, we re-run DeepMNE with different numbers of dimensions for function prediction on yeast dataset and fix other parameters. Figure~\ref{fig:robustness}(a) shows that performance of DeepMNE is stable over a wide range of number of dimensions. In addition, we also test the robustness of DeepMNE to the restart probabilities by varying restart probabilities and fixing other parameters. From the results, we can find that the performance of DeepMNE is stable on different restart probabilities. 

The number of integrated layers may impact the performance of DeepMNE.We tested our method on yeast dataset with different number of layers. DeepMNE can achieves the highest performance when the number of layers is 5 (see Table 5).

\begin{table}[ht]
\caption{Results with different numbers of layers} 
\centering 
\begin{tabular}{c c c c c} 
\hline\hline 
 & Layer-3 & Layer-5 & Layer-7 & Layer-10\\ [0.5ex] 
\hline 
Accuracy & 0.8203 & 0.8248 & 0.8086 & 0.8095 \\
micro-F1 & 0.5977 & 0.5994 & 0.5888 & 0.5906 \\
AUPRC & 0.7366 & 0.7452 & 0.7358 & 0.7346 \\
AUROC & 0.9036 & 0.9097 & 0.9056 & 0.9066 \\ [1ex] 
\hline 
\end{tabular}
\label{table:01} 
\end{table}

\vspace{-5pt}
\section{Conclusions}
Network Embedding, aiming to learn non-linear and low-dimensional feature representation of nodes in networks, has achieved a huge success on many tasks, such as node classification and link prediction. However, current network embedding methods mainly focus on single-network embedding, and few approaches try to learn multi-networks topological information. In this paper, we propose a novel multi-networks embedding algorithm based on semi-supervised autoencoder, termed as DeepMNE. Our approach captures multi-network topological information and takes the correlation among multi-networks into account. We apply our multi-network embedding method on the task of gene function prediction. The experimental results show that DeepMNE outperforms than other state-of-the-art methods and has strong robustness to the number of dimensions and restart probability.


\bibliography{document}
\bibliographystyle{aaai}

\end{document}